\documentclass{ieeeaccess}
\usepackage{cite}
\usepackage{amsmath,amssymb,amsfonts}
\usepackage{graphicx}
\usepackage{textcomp}
\usepackage[T1]{fontenc}
\usepackage[utf8]{inputenc}
\usepackage{placeins}

\DeclareUnicodeCharacter{2009}{\,}

\newlength{\xfigwd}

\graphicspath{{projectImages/}}

\def\BibTeX{{\rm B\kern-.05em{\sc i\kern-.025em b}\kern-.08em
    T\kern-.1667em\lower.7ex\hbox{E}\kern-.125emX}}

\begin{document}
\history{Date of publication xxxx 00, 0000, date of current version xxxx 00, 0000.}
\doi{10.1109/ACCESS.2017.DOI}

\title{Evidence of Phase Transitions in Small Transformer-Based Language Models}

\author{\uppercase{Noah Hong}\authorrefmark{1}, 
\uppercase{Tao Hong}\authorrefmark{2}}

\address[1]{Lynbrook High School, San Jose, CA 95129 USA (e-mail: noahhong01@gmail.com)}
\address[2]{Keysight Technologies, Santa Rosa, CA 95403 USA (e-mail: taohong01@gmail.com)}

\markboth
{Hong \headeretal: Evidence of Phase Transitions in Small Transformer-Based Language Models}
{Hong \headeretal: Evidence of Phase Transitions in Small Transformer-Based Language Models}

\corresp{Corresponding author: Noah Hong (e-mail: noahhong01@gmail.com).}

\begin{abstract}
Phase transitions have been proposed as the origin of emergent abilities in large language models (LLMs), where new capabilities appear abruptly once models surpass critical thresholds of scale. Prior work, such as that of Wei \emph{et al.}, demonstrated these phenomena under model and data scaling, with transitions revealed after applying a log scale to training compute. In this work, we ask three complementary questions: (1) Are phase transitions unique to large models, or can they also be observed in small transformer-based language models? (2) Can such transitions be detected directly in \emph{linear training space}, rather than only after log rescaling? and (3) Can these transitions emerge at early stages of training? To investigate, we train a small GPT-style transformer on a character-level corpus and analyze the evolution of vocabulary usage throughout training. We track the average word length, the number of correct versus incorrect words, and shifts in vocabulary diversity. Building on these measures, we apply Poisson and sub-Poisson statistics to quantify how words connect and reorganize. This combined analysis reveals a distinct transition point during training. Notably, these transitions are not apparent in standard loss or validation curves, but become visible through our vocabulary- and statistics-based probes. Our findings suggest that phase-transition reorganizations are a general feature of language model training, observable even in modest models, detectable directly in linear training space, and occurring surprisingly early as coherence emerges. This perspective provides new insight into the nonlinear dynamics of language model training and underscores the importance of tailored metrics for uncovering phase transition behaviors.
\end{abstract}
\begin{keywords}
Transformer, Phase transition, Emergent abilities, Poisson statistics, KL divergence, Language models
\end{keywords}

\titlepgskip=-15pt
\maketitle
\section{Introduction}

Frontier Large language models (LLMs) have demonstrated striking emergent capabilities in reasoning, coding, and multimodal understanding \cite{radford2018improving, radford2019gpt2, brown2020gpt3, chowdhery2022palm, openai2023gpt4}. These breakthroughs have been widely associated with scaling: as models grow in parameter count, dataset size, and training compute, they appear to acquire qualitatively new abilities. Wei \emph{et al.} characterized such emergent behaviors as \emph{phase transitions}, highlighting that they often arise abruptly at critical thresholds rather than increasing smoothly with scale \cite{wei2022emergent}. 

Despite this progress, most prior studies analyze phase transitions indirectly, requiring billion-parameter models and reporting results only after applying logarithmic rescaling of compute or training steps. As a result, it remains unclear whether phase-transition-like reorganizations can be observed in small, tractable models rather than only in massive LLMs, and raises the question whether such discontinuities can be detected directly on a linear training axis without resorting to log transformations. 

Addressing this gap is important for two reasons. First, small models serve as practical laboratories: they can be trained and inspected with modest resources while still revealing fundamental learning dynamics. Second, if phase transitions are observable directly in raw training steps, this strengthens the case that emergent behaviors are an intrinsic property of neural training, not merely an artifact of evaluation metrics or log scaling. Such insights can inform both interpretability research and the design of more efficient models. 

In this work, we investigate phase-transition dynamics in a small GPT-style transformer (3.6M parameters) trained with character-level tokenization on the Tiny Shakespeare corpus. To detect discontinuities, we introduce a \emph{Poisson-centered diagnostic framework}\cite{hong2023coherent} that tracks dispersion, Kullback–Leibler (KL) divergence, vocabulary composition, and word length. These statistical probes reveal synchronized discontinuities in training that coincide with a transition from random, fragmentary sequences to coherent words. We argue that lexical-level coherence functions as an order parameter for emergent linguistic ability. 

The remainder of this paper is organized as follows. Section II reviews related work on emergent abilities, phase transitions, discontinuous training dynamics, and statistical probes. Section III describes the experimental setup and diagnostic methodology. Section IV presents our findings on synchronized discontinuities. Section V discusses implications, limitations, and directions for future work.

\section{RELATED WORK}
\label{sec:related}

Our investigation of phase transitions in small transformer-based language models draws on three interconnected research traditions: the physics of phase transitions and collective phenomena, the empirical discovery of emergent abilities in large language models, and mechanistic studies of discontinuous learning dynamics. We review each tradition in turn, highlighting how they converge to motivate our central questions about the observability, detectability, and timing of phase-transition-like reorganizations in modest-scale systems.

\subsection{Phase Transitions in Statistical Physics}

The formal study of phase transitions originates in statistical mechanics, where smooth microscopic laws give rise to singular macroscopic behavior. Landau and Lifshitz's canonical treatment establishes the foundational concepts: an order parameter---a macroscopic quantity characterizing the state of the system---changes abruptly as a control parameter crosses a critical threshold \cite{landau2013statistical}. In second-order (continuous) transitions, the order parameter evolves smoothly from zero to finite value, accompanied by diverging fluctuations and correlation lengths. In first-order (discontinuous) transitions, the order parameter jumps discontinuously, reflecting the coexistence and exchange of distinct phases. Crucially, these qualitative reorganizations emerge from collective interactions among many components, not from singularities in the underlying microscopic rules. The water--vapor transition provides an intuitive example: as temperature or pressure varies, the system's density undergoes a sharp, measurable discontinuity despite the smoothness of intermolecular forces.

Tong's modern pedagogical exposition extends this framework through renormalization group (RG) theory and effective field descriptions \cite{tong2011statmech}. RG analysis reveals how critical behavior is governed by fixed points in the space of couplings, organizing transitions into universality classes defined by shared scaling exponents rather than microscopic details. Tong emphasizes that apparent macroscopic discontinuities arise from the amplification of microscopic fluctuations through many-body correlations---a perspective that has proven fruitful when transferred to high-dimensional learning systems. His treatment of effective potentials, in which the system's state concentrates near minima of a coarse-grained free energy, provides a conceptual template later adopted by theorists modeling neural network training as navigation through an effective landscape of competing representations.

Together, Landau--Lifshitz and Tong establish a rigorous vocabulary for describing qualitative reorganizations: order parameters, control parameters, critical thresholds, coexistence regimes, and scaling laws. Although developed for physical systems in thermal equilibrium, this language has migrated to describe abrupt changes in artificial systems, including neural networks, where ``temperature'' may be replaced by training time, learning rate, or model scale, and ``magnetization'' by generalization accuracy, representational coherence, or statistical dispersion.

\subsection{Emergence as a General Principle in Complex Systems}

The conceptual leap from physics to broader contexts was articulated influentially by Anderson, who argued that ``more is different''---that increasing complexity introduces qualitatively new organizational principles not reducible to component-level laws \cite{anderson1972more}. Using examples from superconductivity to molecular biology, Anderson asserted that each level of scale in a hierarchy exhibits emergent properties and demands its own effective theories. This essay laid the philosophical groundwork for studying emergent computation, collective intelligence, and self-organization across disciplines. For artificial intelligence, Anderson's thesis legitimizes the expectation that large-scale neural systems may undergo abrupt reorganizations in capability or structure that cannot be predicted from isolated neuron dynamics or gradient updates.

Building on this foundation, Huberman and Hogg demonstrated that artificial intelligence systems---specifically, symbolic search and constraint-satisfaction algorithms---can exhibit sharp threshold phenomena as control parameters vary \cite{huberman1987phase}. They analyzed how problem difficulty peaks near phase boundaries between solvable and unsolvable regimes, analogous to critical slowing down in physics. Their work established that combinatorial and computational systems, even far from thermodynamic equilibrium, display transition-like behaviors where collective properties change abruptly. Although their focus was symbolic AI rather than deep learning, the conceptual parallel is clear: large computational systems can have critical regions where behavior reorganizes discontinuously.

Forrest extended these ideas into the realm of emergent computation in distributed systems, including early neural networks and evolutionary algorithms \cite{forrest1990emergent}. She argued that computation itself can be an emergent property arising from local interactions that self-organize into global structure. By reviewing cellular automata, associative memories, and genetic algorithms, Forrest highlighted how non-linear coupling of simple rules produces persistent patterns and adaptation. While largely qualitative and pre-dating modern deep learning, her synthesis situates neural training within the broader study of complex adaptive systems, reinforcing the expectation that emergent coherence in language models---such as the abrupt acquisition of multi-character words we observe---can arise from smooth, distributed updates.

\subsection{Phase Transitions in Neural Networks: Early Theoretical Foundations}

The application of phase-transition concepts directly to neural networks began in earnest in the 1990s. Kinzel's review article provided one of the first systematic treatments, analyzing how simple learning systems---perceptrons, Hopfield networks, and spin-glass models---exhibit threshold behavior \cite{kinzel1998phase}. For instance, a perceptron trained on random patterns transitions sharply from an unlearnable to a learnable regime once the ratio of training examples to weights exceeds a critical value. Similarly, associative memory networks undergo order--disorder transitions when capacity limits are exceeded, with retrieval accuracy collapsing abruptly beyond a threshold. Kinzel connected these findings to energy-landscape reorganization and symmetry breaking familiar from condensed-matter physics. His work demonstrated that learning dynamics naturally partition into distinct regimes---ordered (generalizing) and disordered (memorizing or random)---separated by sharp boundaries. While the models studied were low-dimensional and analytically tractable, Kinzel's analysis established that phase-transition language applies meaningfully to neural computation, foreshadowing later efforts to describe deep learning through statistical mechanics.

In the modern era, Bahri and colleagues have synthesized a decade of research applying statistical-mechanical tools to deep neural networks \cite{bahri2023statmech}. Their comprehensive review organizes theory around three axes: the geometry of high-dimensional loss landscapes, the dynamics of signal propagation and criticality (including ``edge-of-chaos'' regimes where networks balance order and chaos), and scaling laws linking architecture, dataset size, and generalization. They describe how random-matrix theory and mean-field approximations reveal phase-transition-like separations between ordered and chaotic signal-flow regimes, and how such transitions influence trainability. Importantly, Bahri et al.\ caution that deep learning operates far from thermodynamic equilibrium; hence, ``phase transition'' serves as a metaphor for qualitative regime changes in optimization dynamics rather than literal thermodynamic phenomena. Nonetheless, the conceptual apparatus---order parameters, critical points, universality---provides a powerful framework for interpreting abrupt shifts in network behavior. For our study, Bahri et al.\ supply rigorous justification for treating discontinuities in dispersion or KL divergence as emergent reorganizations of representation statistics.

Miller, Zhang, and Ganguli have recently proposed that deep networks trained by gradient descent can be modeled as non-equilibrium systems undergoing absorbing-state phase transitions \cite{miller2023absorbing}. They define an order parameter based on the fraction of active degrees of freedom in weight space and demonstrate that training dynamics approach an absorbing state where updates cease, analogous to the frozen phase in directed-percolation universality classes. Empirical scaling exponents measured on feed-forward and recurrent architectures match those predicted by non-equilibrium statistical physics, suggesting universality across network types. While highly theoretical and not linguistically oriented, their work provides formal grounding for describing sharp onsets of regularization---such as the sub-Poisson stabilization we observe---as regime transitions in high-dimensional optimization.

\subsection{Emergent Abilities in Large Language Models}

The empirical discovery of discontinuous capability gains in large-scale language models has catalyzed recent interest in phase-transition analogies. The GPT series documents this progression. Radford et al.\ introduced GPT-1, demonstrating that unsupervised generative pretraining on a large text corpus followed by supervised fine-tuning yields transferable representations that substantially outperform models trained from scratch \cite{radford2018improving}. The key insight was that a single, generic language model can adapt to many downstream tasks with minimal labeled data, suggesting that broad linguistic structure emerges from next-token prediction alone. GPT-2 scaled this paradigm to 1.5 billion parameters and showed that zero-shot multitask performance arises spontaneously without explicit fine-tuning \cite{radford2019gpt2}. Capabilities such as translation, summarization, and question answering appeared organically, motivating the notion that sufficient scale and data diversity induce qualitative leaps in coherence and reasoning.

Brown et al.'s GPT-3, at 175 billion parameters, established scaling as the dominant driver of emergent generalization \cite{brown2020gpt3}. While training loss followed smooth power laws, task performance exhibited abrupt gains in reasoning, arithmetic, and translation once model size exceeded tens of billions of parameters. The paper introduced few-shot, one-shot, and zero-shot evaluation protocols, showing that in-context learning---without gradient updates---becomes viable at scale. GPT-3 thus provided large-scale empirical evidence that internal representations cross thresholds beyond which qualitatively new behaviors emerge, paralleling the discontinuities we observe in miniature systems.

Subsequent frontier models reinforced this pattern. Chowdhery et al.'s PaLM, at 540 billion parameters, demonstrated emergent reasoning at unprecedented scale, with performance curves showing discontinuous jumps in logical and multi-step tasks once critical thresholds were surpassed \cite{chowdhery2022palm}. The authors attributed improvements to both raw scale and dataset diversity, noting that emergent capabilities correlate with cross-domain exposure. Similarly, OpenAI's GPT-4 Technical Report documented marked improvements in reasoning, coding, and multimodal comprehension, with performance on complex tasks again exhibiting threshold-like behavior \cite{openai2023gpt4}. While specific architectural details remain undisclosed, GPT-4's results anchor the empirical end of the scaling continuum, showing that emergent abilities persist and strengthen as models grow.

Wei et al.\ systematically formalized the concept of emergent abilities, defining them as capabilities that cannot be linearly extrapolated from smaller models' performance \cite{wei2022emergent}. By analyzing hundreds of evaluation points across model families---GPT-3, Gopher, Chinchilla, PaLM---they plotted accuracy versus log-scale model size and observed sigmoidal or step-like patterns instead of smooth power-law gains. These discontinuities appeared in arithmetic, translation, reasoning, and word-sense disambiguation. Importantly, Wei et al.\ emphasized that emergent abilities reflect abrupt changes in task performance, not necessarily smooth loss improvements. This observation directly motivated our focus on external behavioral metrics---such as dispersion and vocabulary composition---that reveal reorganizations hidden in standard loss curves. However, Wei et al.\ left the underlying mechanism open, describing emergence primarily as an empirical phenomenon tied to scale.

\subsection{Questioning the Reality of Emergent Abilities}

Schaeffer, Miranda, and Koyejo challenged the interpretation of emergent abilities as evidence of fundamental phase transitions \cite{schaeffer2023mirage}. They argued mathematically and empirically that many step-like curves reported in Wei et al.\ arise from discrete, non-linear evaluation metrics rather than abrupt internal reorganizations. When an underlying performance variable---such as the probability of correct reasoning---increases smoothly with scale, threshold-based metrics like binary accuracy can still produce the illusion of sudden jumps. By re-plotting results using continuous metrics such as cross-entropy or expected loss, Schaeffer et al.\ found mostly smooth scaling laws across model families. They concluded that observed ``emergence'' may often be an artifact of binarized or non-linear task scoring, though they acknowledged that some genuine discontinuities could exist and encouraged mechanistic probes of internal representations.

This critique is foundational for our study. Schaeffer et al.\ demonstrate that not every S-curve or cusp in task performance implies a physical-like phase transition, underscoring the necessity of continuous, internal diagnostics rather than external accuracy thresholds. Our use of Poisson and sub-Poisson statistics, KL divergence, and vocabulary composition directly addresses this concern: these are continuous measures of linguistic structure that evolve throughout training, independent of arbitrary task cutoffs. By detecting synchronized discontinuities across multiple continuous probes---dispersion, KL divergence, word length, vocabulary dynamics---we provide converging evidence for a genuine reorganization rather than a metric artifact. In this sense, our approach heeds Schaeffer et al.'s call for mechanistic, continuous analysis while still identifying discontinuous behavior.

\subsection{Grokking and Discontinuous Training Dynamics}

The phenomenon of grokking, introduced by Power et al., provides the most direct small-scale analogue to our findings \cite{power2022grokking}. Training small transformers and MLPs on modular-arithmetic datasets, Power et al.\ observed that models first overfit---achieving near-zero training loss but random test accuracy---and then, after many additional epochs, suddenly transition to perfect generalization. This delayed, abrupt reorganization occurs despite smooth loss curves and depends on regularization strength (weight decay) and dataset diversity. Visualizations of embedding geometry reveal that during grokking, learned features rotate and align with the true algorithmic structure, suggesting the model moves from a memorization basin to a generalization basin in weight space. Grokking thus establishes that neural networks, even at modest scale, can remain in a long plateau before abruptly reorganizing internally---a discontinuity not captured by training loss alone.

Agarwal, Morcos, and Krishnamurthy extended this work through mechanistic interpretability, applying singular value decomposition, neuron activation analysis, and embedding alignment to track how grokking unfolds inside the network \cite{agarwal2023progress}. They introduced quantitative progress measures capturing the degree of algorithmic abstraction in weights and representations. Their key finding is that even when external accuracy remains flat, internal circuits gradually form compositional structures; once these structures cross a stability threshold, generalization jumps. This threshold coincides with sudden drops in representational entropy and changes in gradient flow, supporting the view that grokking is an internal reorganization event, not gradual refinement. For our work, Agarwal et al.\ provide a mechanistic complement to our external metrics: both reveal latent buildup followed by rapid consolidation. The parallel is direct---our dispersion and KL diagnostics capture analogous reorganizations in linguistic structure, where fragments coalesce into coherent words at a critical training epoch.

Dolev et al.\ documented similar discontinuities in neural machine translation, where performance on morphologically rich and low-resource languages leaps suddenly once model or dataset size crosses a threshold \cite{dolev2023translation}. Using Transformer architectures trained on WMT datasets, they observed that while translation quality grows smoothly at small scales, it jumps sharply beyond a critical point. They attributed the discontinuity to improved contextual representation and capacity to memorize rare-word alignments, terming these ``critical advantages'' of scaling. Although Dolev et al.\ did not analyze internal representation dynamics or apply statistical probes, their empirical curves parallel the abrupt lexical consolidation we observe, reinforcing the idea that neural sequence models can display threshold-like improvements in linguistic coherence.

\subsection{Theoretical Foundations: Grokking as a First-Order Phase Transition}

Rubin, Seroussi, and Ringel provided the first formal statistical-physics model of grokking, directly linking training dynamics to phase transitions \cite{rubin2024grokking}. They studied a two-layer teacher--student network trained on modular addition and derived an effective potential describing the system's macroscopic state. The potential exhibits two competing minima: one corresponding to memorization (fitting training data without generalization) and another to generalization (capturing the underlying rule). As regularization strength or training time increases, the minima exchange stability, producing a discontinuous jump in the order parameter---identifying grokking as a first-order phase transition. Rubin et al.\ supported this mapping through analytical mean-field approximations and numerical simulations, demonstrating hysteresis and mixed phases characteristic of first-order transitions. They also identified intermediate regimes with Gaussian mixture features, matching empirical embedding structures seen in earlier grokking studies.

This theoretical work establishes the strongest justification for describing discontinuities in neural training as genuine phase transitions. Rubin et al.\ demonstrate rigorously that smooth local updates---gradient descent steps---can yield abrupt, global changes in representational order once competing learning modes exchange stability. Although their model applies to simplified two-layer networks and algorithmic tasks, the conceptual framework extends naturally to our setting. The dispersion flip we observe---correct words transitioning from near-Poisson to sub-Poisson while incorrect words move from sub-Poisson to Poisson-like---can be interpreted as the system crossing a critical threshold where the ``generalization minimum'' (coherent word usage) overtakes the ``memorization minimum'' (fragmentary output). In this view, our Poisson-centered diagnostics serve as order parameters tracking which representational mode dominates as training progresses.

\subsection{Positioning Our Work}

Against this backdrop, our study occupies a distinct and underexplored position. Most empirical investigations of emergent abilities (Wei et al., Brown et al., Chowdhery et al.) focus on billion-parameter models evaluated after training, often requiring logarithmic rescaling of compute or model size to reveal discontinuities. In contrast, we investigate a small, 3.6M-parameter transformer and track its evolution directly along the linear training axis, demonstrating that phase-transition-like reorganizations are observable in modest systems without log-scaling. Grokking research (Power et al., Agarwal et al.) provides the closest analogue---small models, abrupt reorganization---but focuses on algorithmic datasets and generalization accuracy. We extend this line to linguistic tasks, introducing Poisson-based statistical probes that detect discontinuities in vocabulary structure, word length, and dispersion rather than task accuracy alone.

Theoretically, we leverage the frameworks of Bahri et al., Miller et al., and especially Rubin et al.\ to interpret our findings. The dispersion flip---correct words shifting from near-Poisson to sub-Poisson, incorrect words from sub-Poisson to Poisson-like---mirrors the competition between memorization and generalization minima in Rubin et al.'s effective potential. Our KL divergence and vocabulary dynamics corroborate this interpretation, showing synchronized cusps and collapses coinciding with the onset of multi-character word coherence. Methodologically, we heed Schaeffer et al.'s caution by employing continuous, internal metrics rather than binary task thresholds, thereby avoiding metric-induced artifacts while still identifying genuine discontinuities.

In sum, our contribution is threefold: (1) we demonstrate that phase-transition-like reorganizations are not unique to large language models but arise in small transformers trained on character-level linguistic data; (2) we show that such transitions can be detected directly in linear training space using Poisson-centered diagnostics, without requiring log-scaling or massive compute; and (3) we establish that these transitions occur surprisingly early in training as lexical coherence emerges, with word-level structure serving as a natural order parameter. By bridging statistical physics, emergent abilities research, and small-model training dynamics, we provide both empirical evidence and interpretive tools for understanding nonlinear reorganizations in neural language learning.

\section{Methods}

\subsection{Model and Data}
We trained a compact character-level transformer designed to resemble the basic architecture of larger language models while remaining computationally controllable. 
The architecture consists of an embedding dimension of 192, 8 transformer layers, and 6 attention heads (approximately 3.6M parameters) with a context length of 128 characters. 
The model was trained on the Tiny Shakespeare corpus, which contains $\sim$ 1.1M character tokens that span 65 unique characters.  

To assess robustness, we ran 5 independent seeds for 0--600 epochs. 
At each checkpoint, we sampled 30{,}000 tokens using a fixed decoding configuration (temperature $T=1.0$, greedy top-1 unless otherwise noted). 

\subsection{Segmentation and Correctness Labeling}
Generated text was segmented into words using whitespace and punctuation boundaries.  
We then classified each unit as either:
\begin{itemize}
    \item \textbf{Correct:} if it appears in the corpus vocabulary.
    \item \textbf{Incorrect:} if it does not appear in the corpus vocabulary.
\end{itemize}
This operational split provides a way to separate the legitimate vocabulary from exploratory or erroneous output. 
Although it excludes words beyond the training distribution, this operational definition still captures the model's progression toward vocabulary-level structure, thereby offering a practical basis for experimental analysis.
\subsection{Vocabulary Growth and Word Length}
To capture the linguistic impact of the transition, we track two complementary measures that quantify the model's progression from fragmentary output to compositional word formation.

\textbf{Unique vocabulary counts} ($V_{\text{uniq,correct}}$ and $V_{\text{uniq,incorrect}}$) measure the diversity of word types generated in each class. The unique correct vocabulary reflects the model's expanding repertoire of legitimate forms, while the unique incorrect vocabulary tracks the proliferation and eventual collapse of erroneous patterns. By monitoring both separately, we can identify when the model shifts from exploring fragmentary errors to consolidating valid words. A peak in incorrect vocabulary diversity followed by collapse signals that the model has exhausted exploratory combinations and committed to structured patterns.

\textbf{Average word length} ($\bar{L}$) provides a direct measure of compositional complexity:
\begin{equation}
\bar{L} = \frac{1}{N_{\text{correct}}}\sum_{w \in \text{correct}} \mathrm{len}(w).
\end{equation}
Early in training, when the model primarily generates single-character outputs or short two-letter fragments, $\bar{L}$ remains near 1.5 characters. As the model learns to compose longer, multi-character words, $\bar{L}$ increases. A sharp, S-shaped rise in average word length signals a qualitative shift from character-level fragmentation to word-level composition. This metric serves as a natural linguistic correlate of the dispersion transition: sub-Poisson regularity in correct-word usage emerges precisely when the model begins producing stable, multi-character forms.

Together, these measures reveal how the system reorganizes from producing short, error-prone fragments to longer coherent words. The synchronized dynamics—incorrect vocabulary collapse, correct vocabulary acceleration, and word-length increase—provide converging evidence that the statistical reorganization captured by dispersion and KL divergence manifests as a tangible linguistic restructuring.

\subsection{Poisson Baseline, Dispersion, and Snapshots}
To test whether words appear randomly or in a structured way, we divided generated text into non-overlapping windows of $W=21$ words. 
If word occurrences were independent random events, the number of words $N$ per window would follow a Poisson distribution:
\begin{equation}
P(N=n;\lambda) = \frac{\lambda^n e^{-\lambda}}{n!}, 
\qquad \mathrm{Var}[N] = \lambda,
\end{equation}
with mean $\lambda$ equal to the variance.  

Deviations from this Poisson baseline are summarized by the 
\textbf{index of dispersion}:
\begin{equation}
D = \frac{\sigma^2}{\mu},
\end{equation}
where $\mu$ and $\sigma^2$ are the mean and variance of the count distribution.  

\begin{itemize}
    \item $D=1$: \textbf{Poissonian} — events occur randomly and independently, with variance equal to the mean. This represents the natural benchmark of uncorrelated randomness.
    \item $D<1$: \textbf{Sub-Poissonian} — variance is suppressed relative to the mean. Events are more evenly spaced than chance would predict, signaling hidden order or regularity. In physics, this regime is associated with quantum antibunching of photons.
    \item $D>1$: \textbf{Super-Poissonian} — variance exceeds the mean. Events occur in bursts or clusters, as seen in chaotic or heavy-tailed processes.
\end{itemize}

In our linguistic setting, these regimes carry intuitive interpretations.  
\begin{itemize}
    \item \textbf{Correct words} migrate from near-Poisson to sub-Poisson as training progresses, reflecting a shift from rare, scattered successes to stable, regular usage patterns.  
    \item \textbf{Incorrect words} begin in a repetitive, variance-suppressed regime (sub-Poisson) dominated by short, recurring fragments like ``th'' or ``yo,'' then disperse randomly as errors diminish, drifting toward $D \approx 1$ as they become sparse, independent noise.  
\end{itemize}

\begin{figure*}[!t]
\centering
\includegraphics[width=0.8\textwidth]{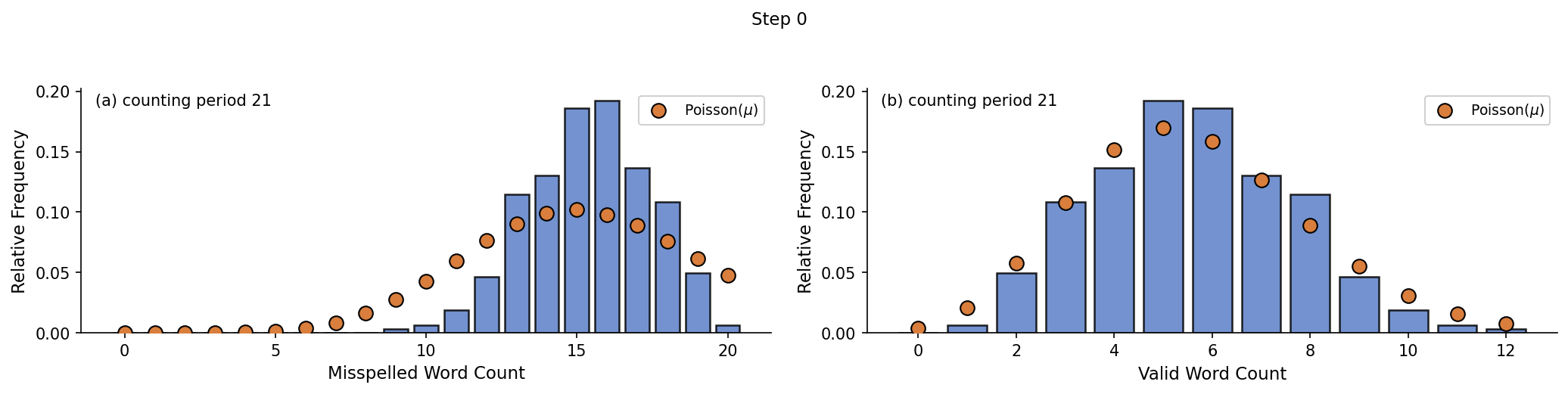}
\caption{\textbf{Step 0 (Early training).} Side-by-side histograms of windowed word counts. \textbf{Left:} Incorrect words show sub-Poisson clustering with narrow distributions, reflecting repetitive short fragments. \textbf{Right:} Correct words are extremely sparse and scattered, displaying near-Poisson randomness.}
\label{fig:hist_step0}
\end{figure*}

\begin{figure*}[!t]
\centering
\includegraphics[width=0.8\textwidth]{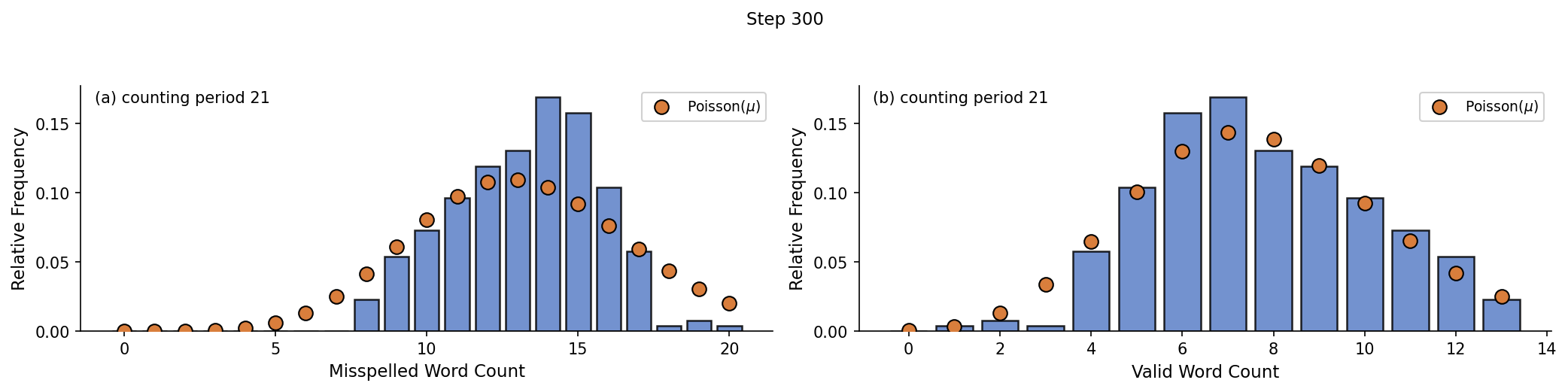}
\caption{\textbf{Step 300 (Mid-training, approaching transition).} \textbf{Left:} Incorrect words maintain sub-Poisson clustering with narrow, variance-suppressed distributions as repetitive fragments persist. \textbf{Right:} Correct words begin showing nascent structure but remain scattered and near-Poisson.}
\label{fig:hist_step300}
\end{figure*}

\begin{figure*}[!t]
\centering
\includegraphics[width=0.8\textwidth]{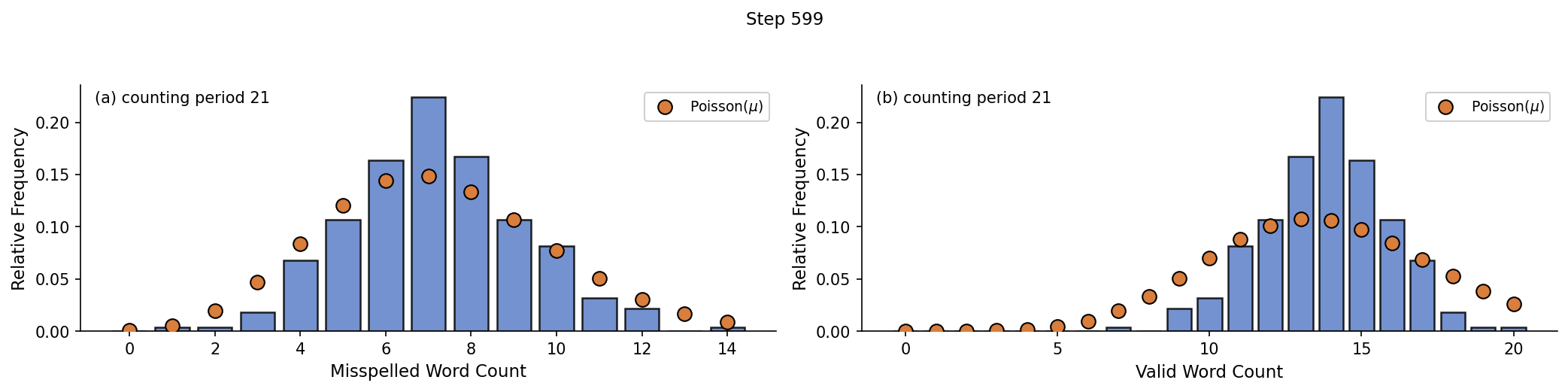}
\caption{\textbf{Step 599 (Post-transition).} The dispersion flip is complete. \textbf{Left:} Incorrect words have broadened to near-Poisson distributions, becoming sparse, independent residual errors that occur randomly. \textbf{Right:} Correct words have contracted into tight, sub-Poisson distributions, reflecting regular, structured usage.}
\label{fig:hist_step599}
\end{figure*}

To illustrate these dynamics, Figs.~\ref{fig:hist_step0}--\ref{fig:hist_step599} show side-by-side histograms of incorrect (left) and correct (right) word counts at representative checkpoints. Each panel displays the empirical distribution (blue bars) overlaid with the fitted Poisson baseline (orange dots).

\textbf{Step 0 (Early training):} Incorrect words show sub-Poisson clustering, with narrow distributions reflecting repetitive short fragments. Correct words are extremely sparse and scattered, displaying near-Poisson randomness with very few successful multi-character forms.

\textbf{Step 300 (Mid-training, approaching transition):} Incorrect words maintain sub-Poisson clustering with narrow, variance-suppressed distributions as repetitive fragments persist. Correct words begin to show emerging structure but remain scattered and near-Poisson, indicating the model is starting to discover valid patterns but usage remains irregular.

\textbf{Step 599 (Late training, post-transition):} The dispersion flip is complete. Incorrect words have broadened to near-Poisson distributions, consistent with sparse, independent residual errors that occur randomly rather than in repetitive clusters. Correct words have contracted into tight, sub-Poisson distributions with suppressed variance, reflecting regular, structured usage where coherent words appear predictably and consistently.

This visual progression directly illustrates the dispersion flip that defines the transition: incorrect words evolve from sub-Poisson (repetitive clustered fragments like ``th, th, th'') to Poisson (sparse random errors), while correct words evolve from near-Poisson (scattered rare successes) to sub-Poisson (regular coherent usage).

\FloatBarrier

\subsection{KL Divergence from Poisson}
While dispersion captures deviations through the variance-to-mean ratio, it cannot detect all distributional changes. To complement this analysis, we compute the Kullback--Leibler (KL) divergence between the empirical histogram $\hat{p}(n)$ of windowed counts and a fitted Poisson distribution with mean $\hat{\lambda}$:
\begin{equation}
D_{\mathrm{KL}}(\hat{p}\,\|\,\mathrm{Pois}(\hat{\lambda})) 
= \sum_{n} \hat{p}(n)\,\log\frac{\hat{p}(n)}{P(n;\hat{\lambda})}.
\end{equation}
Here $D_{\mathrm{KL}}=0$ only if the empirical distribution matches the Poisson baseline exactly. 
KL divergence measures the full distributional deviation from Poisson, providing sensitivity to higher-order statistical structure beyond variance alone. 
Together, dispersion and KL divergence form complementary probes: dispersion directly quantifies the regime (Poisson vs.\ sub-Poisson), while KL divergence quantifies the magnitude of deviation. 
The synchronization of cusps in both metrics strengthens evidence for genuine reorganization rather than metric artifacts.

\subsection{Estimation and Visualization}
All metrics are averaged across 5 seeds, with shaded error bands (where plotted) denoting $\pm$ one standard deviation.  
Histograms are shown at representative early and late checkpoints.  
All decoding parameters are fixed unless otherwise noted. 

\section{Results}

\subsection{Overview of Discontinuities}
Across all five seeds, every external metric exhibits a coordinated, transition-like reorganization
that is not visible in the smooth training/validation losses. In the same epoch band, we observe:
(i) synchronized cusps in both dispersion and KL divergence; 
(ii) a temporary reversion of correct-word dispersion toward Poisson before stabilizing to sub-Poisson; 
(iii) a rise--then--collapse of the unique \emph{incorrect} vocabulary alongside steady growth of the unique \emph{correct} vocabulary; 
and (iv) an S-curve jump in average correct-word length from $\sim$1.5 to $\sim$2.5 characters. 
Qualitatively, the model transitions from producing short fragments and errors to composing multi-character, coherent words.
The convergence of multiple independent, continuous metrics—not just one—supports the interpretation of a genuine phase-transition-like reorganization rather than a gradual drift or metric artifact.

\subsection{Word-Frequency Snapshots at 150, 250, 350, and 500}
\begin{figure}[!htbp]
\centering
\includegraphics[width=0.85\linewidth]{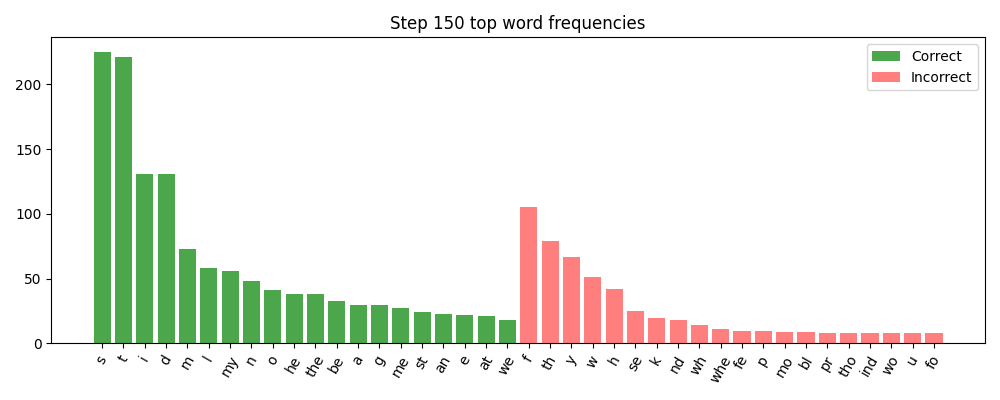}
\caption{\textbf{Step 150 (Pre-transition).} Top generated words are dominated by short fragments and errors (red bars). Single characters and malformed tokens account for most outputs. The model has not yet learned multi-character compositional structure.}
\label{fig:step150}
\end{figure}

\begin{figure}[!htbp]
\centering
\includegraphics[width=0.85\linewidth]{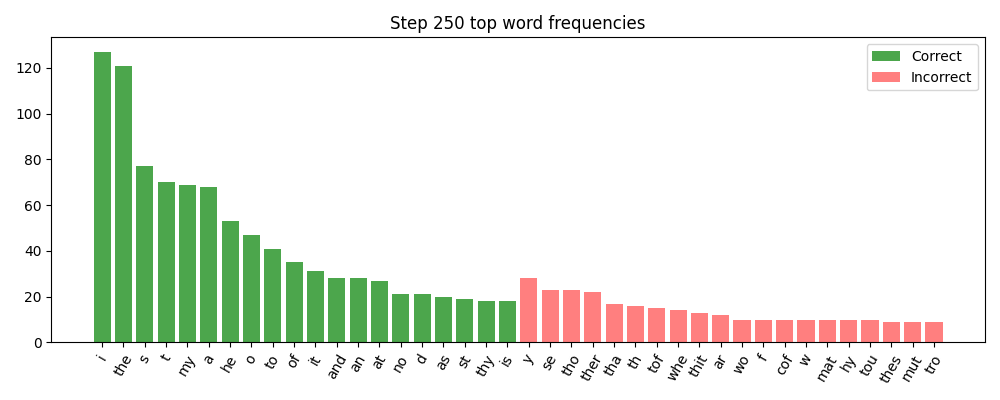}
\caption{\textbf{Step 250 (Transition onset).} Early coherent words (\emph{``to''}, \emph{``is''}, \emph{``my''}) appear alongside persistent errors. The model is in an exploratory regime where valid patterns coexist with fragmentary noise.}
\label{fig:step250}
\end{figure}

\begin{figure}[!htbp]
\centering
\includegraphics[width=0.85\linewidth]{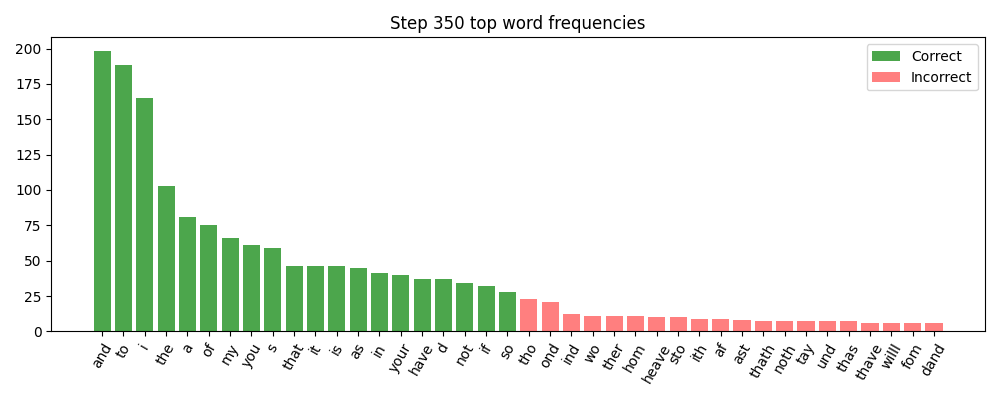}
\caption{\textbf{Step 350 (Transition band).} Mid-length words (\emph{``you''}, \emph{``the''}, \emph{``and''}) emerge decisively. Error diversity collapses. This checkpoint aligns with the synchronized cusps in dispersion and KL divergence.}
\label{fig:step350}
\end{figure}

\begin{figure}[!htbp]
\centering
\includegraphics[width=0.85\linewidth]{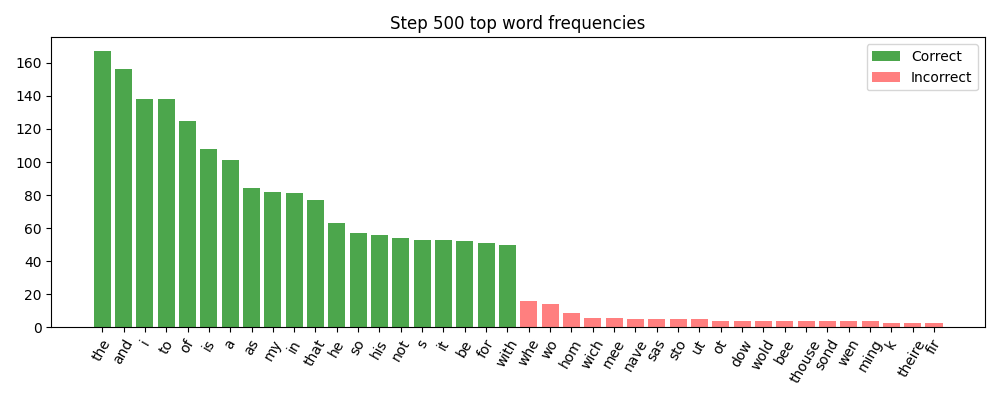}
\caption{\textbf{Step 500 (Post-transition).} Coherent words dominate with stable frequency patterns. High-frequency function words are consistent; errors are sparse. The vocabulary has consolidated around legitimate forms.}
\label{fig:step500}
\end{figure}

To visualize the lexical reorganization concretely, we examine frequency distributions of the top generated words at four representative checkpoints spanning the transition (Figs.~\ref{fig:step150}--\ref{fig:step500}). These snapshots reveal the progression from fragmentary, error-dominated output to coherent, structured vocabulary.

\textbf{Step 150 (Pre-transition):} Generated text is dominated by short fragments and malformed tokens. The most frequent outputs are single characters or defective two-letter combinations that do not correspond to valid English words. Error tokens (shown in red) outnumber correct words, and even the correct tokens tend to be single-character artifacts like punctuation or isolated letters. The model has not yet discovered multi-character compositional structure.

\textbf{Step 250 (Transition onset):} Early coherent words begin to appear alongside persistent errors. Legitimate two- and three-letter words such as \emph{``to''}, \emph{``is''}, and \emph{``my''} emerge with increasing frequency, but the distribution remains mixed. Fragmentary errors still account for a substantial fraction of outputs, indicating the model is in an exploratory phase where valid patterns coexist with residual noise. The balance between correct and incorrect vocabulary is shifting but not yet stable.

\textbf{Step 350 (Transition band):} Mid-length words such as \emph{``you''}, \emph{``the''}, and \emph{``and''} emerge decisively and stabilize at the top of the frequency distribution. Error tokens collapse in both diversity and frequency, now appearing primarily as low-frequency outliers. The shift from single-character fragments to multi-character words is nearly complete. This checkpoint corresponds precisely to the epochs where dispersion and KL divergence exhibit synchronized cusps, confirming that the statistical reorganization manifests as a lexical restructuring.

\textbf{Step 500 (Post-transition):} Coherent words dominate the distribution, and their relative frequencies have settled into stable patterns reflecting the underlying corpus statistics. High-frequency function words (\emph{``the''}, \emph{``and''}, \emph{``you''}) appear consistently, while error tokens are sparse and relegated to the tail. The vocabulary has consolidated around legitimate forms, and word lengths reflect compositional structure rather than random character sequences.

Together, these snapshots provide concrete lexical evidence for the exploratory $\to$ consolidating shift captured by our Poisson-based metrics. The visual progression from fragment-dominated to word-dominated output directly parallels the dispersion flip (incorrect sub-Poisson $\to$ Poisson, correct near-Poisson $\to$ sub-Poisson) and the KL divergence cusp, demonstrating that the statistical reorganization has clear linguistic correlates.

\FloatBarrier

\subsection{Average Word Length (Correct Words)}
\begin{figure}[!htbp]
\centering
\includegraphics[width=0.85\linewidth]{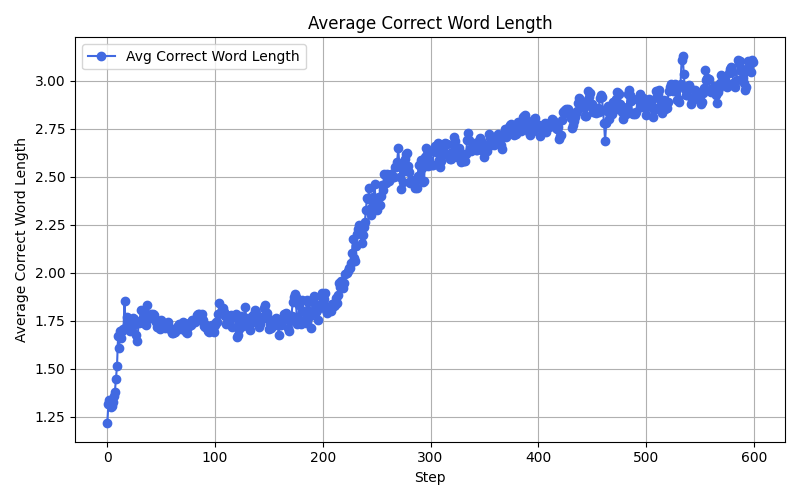}
\caption{Average correct-word length vs.\ epoch. The S-curve jump aligns with the transition band identified by both dispersion and KL divergence.}
\label{fig:avg_len}
\end{figure}

Average correct-word length provides a direct linguistic indicator (Fig.~\ref{fig:avg_len}). After a long, flat phase near single-character outputs ($\sim$1.5), the mean jumps in an S-curve to $\sim$2.5 characters in the same transition band highlighted later by dispersion and KL divergence. The sharp increase signals the onset of composing multi-character words and is consistent with the subsequent sub-Poisson regularity of correct usage.

\FloatBarrier

\subsection{Unique Vocabulary Dynamics}
\begin{figure}[!htbp]
\centering
\includegraphics[width=0.85\linewidth]{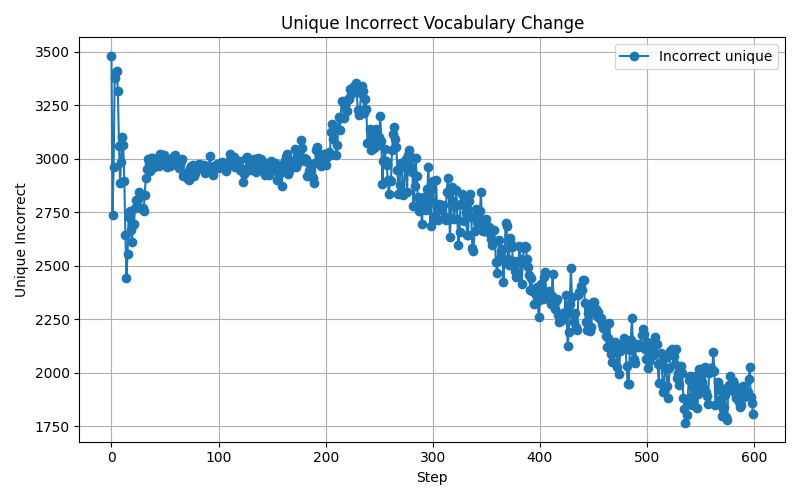}
\caption{Unique incorrect vocabulary: plateau $\to$ peak near step 250 $\to$ collapse. The peak coincides with the transition band.}
\label{fig:uniq_incorrect}
\end{figure}

Unique-type dynamics separate exploration from consolidation. The \emph{incorrect} vocabulary (Fig.~\ref{fig:uniq_incorrect}) exhibits a plateau (steps 0--200), a transient \emph{peak} near step 250 (fragment proliferation), followed by a sustained \emph{collapse} as erroneous forms are abandoned. This peak coincides precisely with the transition band identified by dispersion and KL divergence.

\begin{figure}[!htbp]
\centering
\includegraphics[width=0.85\linewidth]{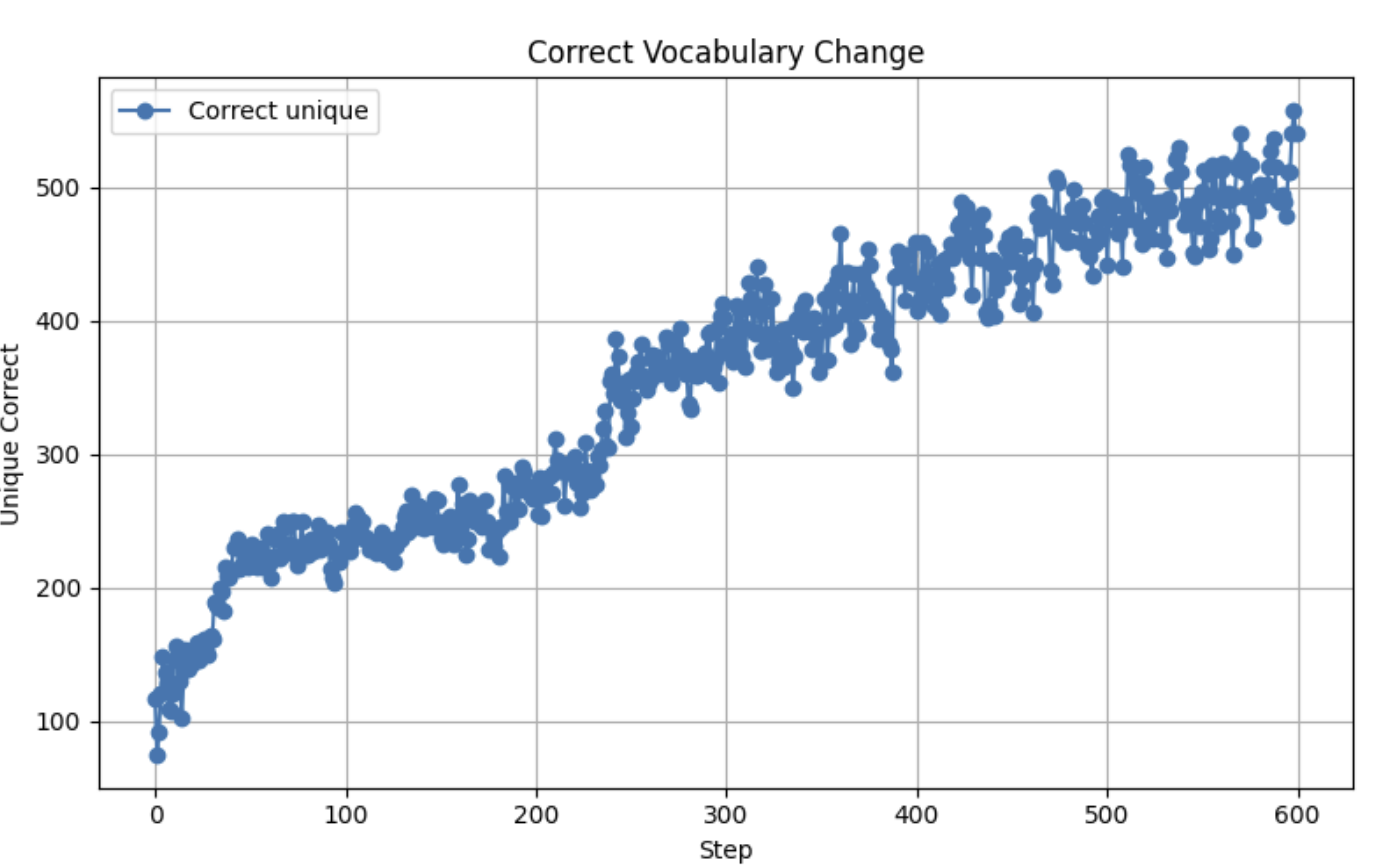}
\caption{Unique correct vocabulary: multi-phase growth with acceleration during the transition (steps 250--400) aligning with dispersion and KL divergence cusps.}
\label{fig:uniq_correct}
\end{figure}

The \emph{correct} vocabulary (Fig.~\ref{fig:uniq_correct}) displays multi-phase growth: rapid initial accumulation (steps 0--100), deceleration and plateau (steps 100--250), sharp acceleration during the transition (steps 250--400), and continued growth post-transition. The acceleration phase aligns with the dispersion and KL cusps, indicating that the statistical reorganization manifests as expansion of the legitimate vocabulary.

The push--pull dynamics—incorrect vocabulary peaks as correct vocabulary accelerates—suggest that fragment proliferation precedes word consolidation: the model explores erroneous combinations before committing to valid forms.

\FloatBarrier

\subsection{Case Study: Prefix Formation and Acquisition of ``you''}
\begin{figure}[!htbp]
\centering
\includegraphics[width=0.95\linewidth]{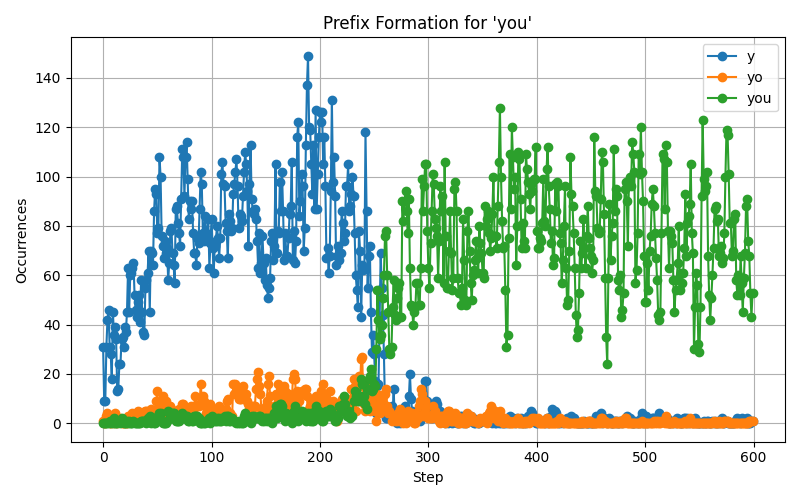}
\setlength{\xfigwd}{0.95\linewidth}
\caption{Prefix formation for ``you'': counts of \emph{y}, \emph{yo}, and \emph{you} across training steps. Near the transition, the full word \emph{you} rises sharply while prefix counts decline, indicating consolidation from fragments into a multi-character word.}
\label{fig:you_prefix}
\end{figure}

To test whether the transition corresponds to composing longer words from shorter fragments, we track the evolution of the word ``you'' and its prefixes. In Fig.~\ref{fig:you_prefix}, counts of \emph{y}, \emph{yo}, and \emph{you} are plotted across steps. Early training is dominated by single-letter \emph{y}; as the transition approaches, \emph{yo} appears transiently; at the transition, the full word \emph{you} rises sharply while \emph{y} and \emph{yo} decline. This pattern is consistent with consolidation of fragments into a stable three-letter form. Fig.~\ref{fig:you_prefix} also shows the learning curve for \emph{you} itself: an S-shaped rise precisely in the band where dispersion and KL divergence display cusps. In Poisson terms, the temporary reversion of correct-word dispersion toward $D\!\approx\!1$ can be interpreted as a brief exploratory phase during recombination; post-transition, correct usage settles into sub-Poisson regularity while residual errors are Poisson-like.

\FloatBarrier

\subsection{KL Divergence to Poisson Baseline}
KL divergence provides an independent measure of how much the empirical word-count distributions deviate from the Poisson baseline. We track correct and incorrect words separately to reveal the divergence--convergence duality that characterizes the transition.

\begin{figure}[!htbp]
\centering
\includegraphics[width=0.85\linewidth]{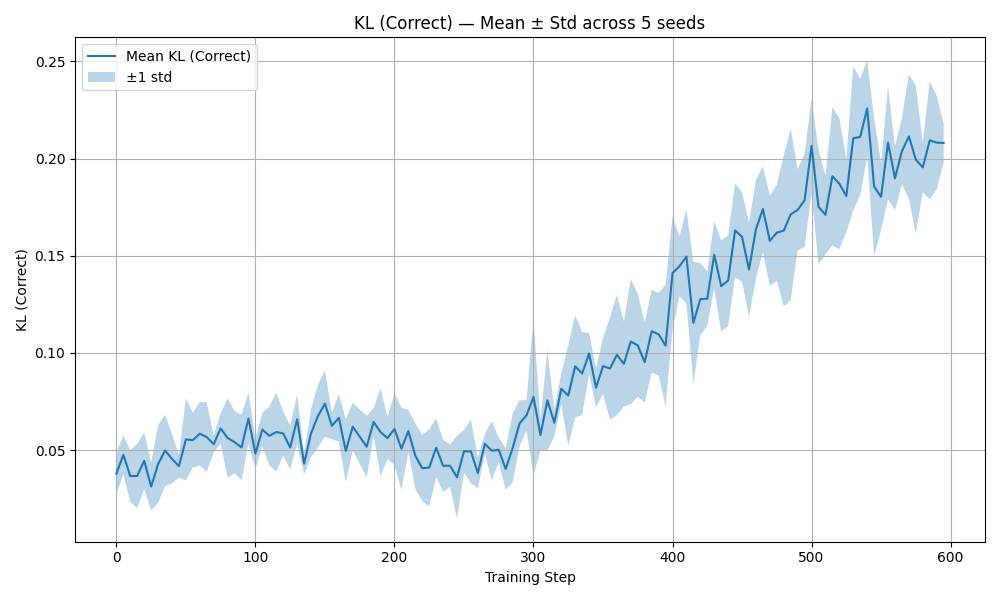}
\caption{\textbf{KL divergence for correct words} to a fitted Poisson baseline across epochs (mean $\pm$ s.d.). A pronounced cusp appears at epochs 230--250 as the model transitions from sparse, near-Poisson usage to structured, sub-Poisson regularity. Post-transition, KL remains elevated, reflecting stable non-Poisson structure.}
\label{fig:kl_correct}
\end{figure}

\textbf{Correct words} (Fig.~\ref{fig:kl_correct}) exhibit a pronounced cusp: KL divergence remains low during the early exploratory regime, then rises sharply at the transition (epochs 230--250) as regular usage patterns emerge. Post-transition, KL divergence remains elevated, reflecting that correct-word distributions stabilize in a structured, non-Poisson regime. The cusp indicates the moment when the model transitions from sparse, random correct-word usage to systematic, sub-Poisson regularity. Notably, KL divergence captures distributional features beyond variance: even when dispersion temporarily returns to $D{\approx}1$ during the exploratory phase, KL divergence remains elevated, indicating that the distribution shape differs from true Poisson randomness.

\begin{figure}[!htbp]
\centering
\includegraphics[width=0.85\linewidth]{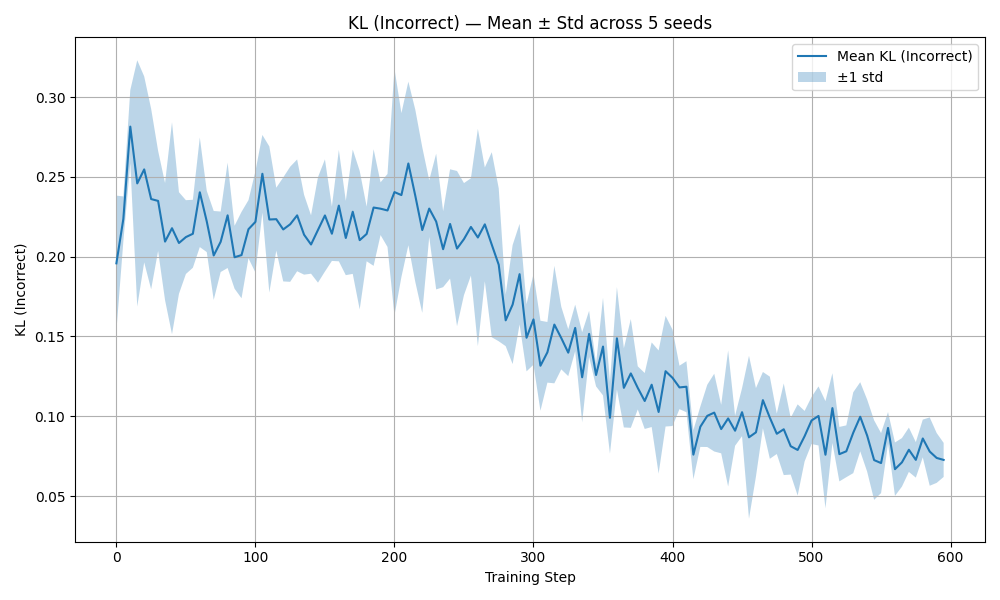}
\caption{\textbf{KL divergence for incorrect words} to a fitted Poisson baseline across epochs (mean $\pm$ s.d.). Initially elevated due to sub-Poisson clustering of fragments, KL converges toward zero as errors become sparse, independent residual noise. The convergence coincides with the correct-word cusp (Fig.~\ref{fig:kl_correct}).}
\label{fig:kl_incorrect}
\end{figure}

\textbf{Incorrect words} (Fig.~\ref{fig:kl_incorrect}) display the opposite pattern: KL divergence is initially elevated (reflecting the sub-Poisson clustering of repetitive fragments), then converges toward zero as errors become sparse and independent. Post-transition, incorrect words approach the Poisson baseline, consistent with their role as rare, uncorrelated residual noise. The convergence of incorrect-word KL toward zero coincides precisely with the collapse of the incorrect vocabulary and the cusp in correct-word KL.

This divergence--convergence duality occurs in the same critical epoch band (230--250) identified by dispersion and word-length metrics. The fact that KL divergence and dispersion exhibit synchronized cusps—yet measure different statistical properties (full distributional form versus variance structure)—demonstrates that the transition is captured by multiple independent probes. Their simultaneous reorganization provides converging evidence for a genuine phase-transition-like event rather than a metric artifact.

\FloatBarrier

\subsection{Index of Dispersion (Fano Factor)}
The index of dispersion ($D{=}\sigma^2/\mu$) acts as our primary order parameter, quantifying whether word occurrences follow Poissonian randomness ($D{=}1$), sub-Poissonian regularity ($D{<}1$), or super-Poissonian clustering ($D{>}1$). We analyze correct and incorrect words separately to reveal the regime flip that defines the transition.

\begin{figure}[!htbp]
\centering
\includegraphics[width=0.85\linewidth]{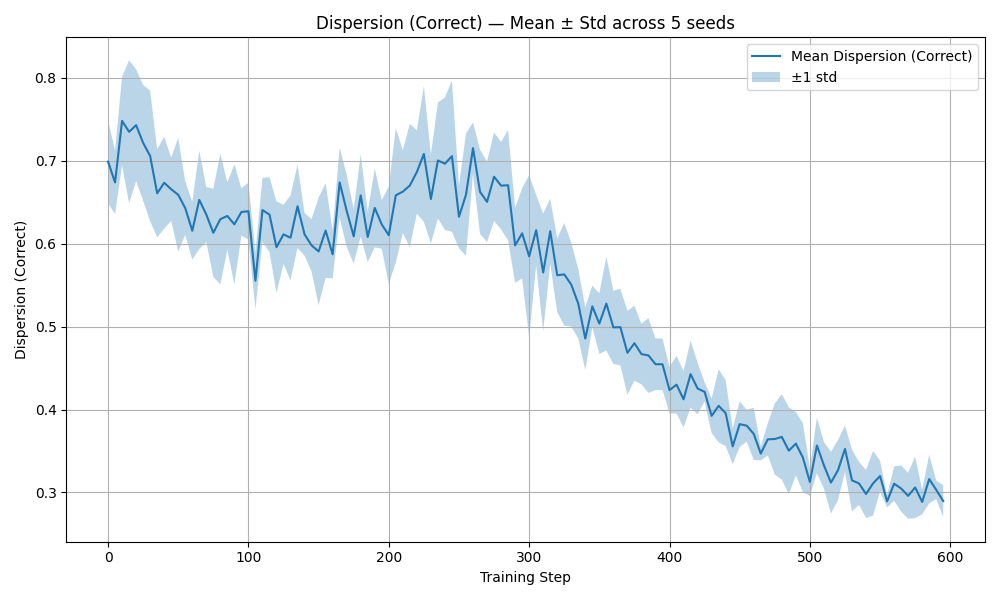}
\caption{\textbf{Index of Dispersion for correct words} ($D=\sigma^2/\mu$) across epochs (mean $\pm$ s.d.). Begins near Poisson ($D{\approx}1$), exhibits a temporary reversion to Poisson at the transition (epochs 230--250), then stabilizes in a sub-Poisson regime ($D{<}1$), indicating regular, structured usage.}
\label{fig:dispersion_correct}
\end{figure}

\textbf{Correct words} (Fig.~\ref{fig:dispersion_correct}) begin near Poisson ($D{\approx}1$) during early training, reflecting sparse, scattered usage. At the transition (epochs 230--250), dispersion exhibits a temporary dip toward $D{\approx}1$ before stabilizing below unity ($D{<}1$) post-transition. This sub-Poisson regime indicates that correct words are generated with regular, evenly-spaced patterns rather than random bursts. The temporary return to Poisson during the transition suggests a brief exploratory phase where the model reorganizes its word-generation strategy before committing to structured usage. This transient precisely matches the KL divergence cusp and the histogram flip shown in the Methods section.

\begin{figure}[!htbp]
\centering
\includegraphics[width=0.85\linewidth]{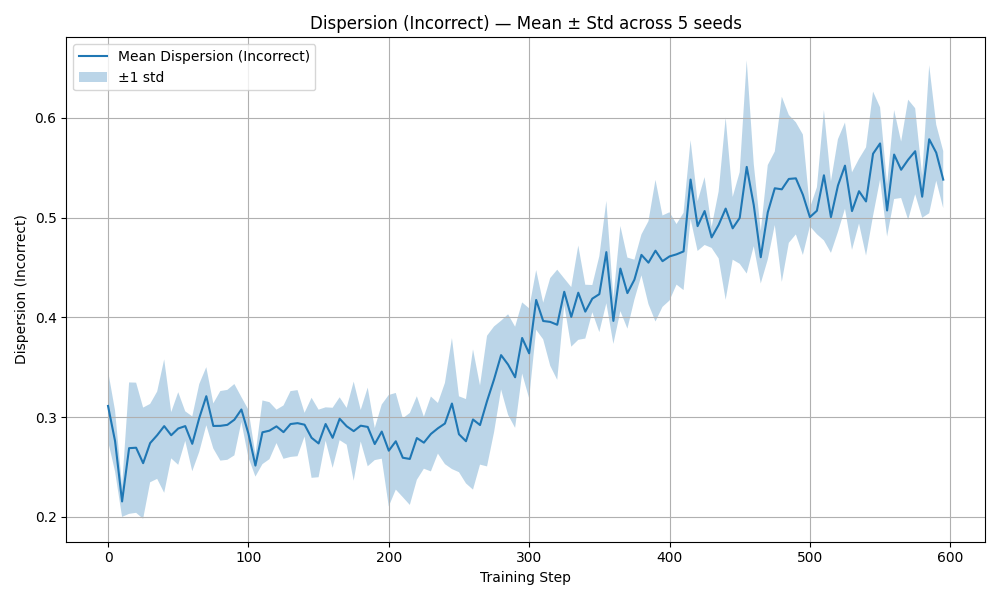}
\caption{\textbf{Index of Dispersion for incorrect words} ($D=\sigma^2/\mu$) across epochs (mean $\pm$ s.d.). Begins in a sub-Poisson regime ($D{<}1$) reflecting repetitive fragments, then transitions toward Poisson ($D{\approx}1$) as errors become sparse and independent. The shift mirrors the correct-word transition (Fig.~\ref{fig:dispersion_correct}).}
\label{fig:dispersion_incorrect}
\end{figure}

\textbf{Incorrect words} (Fig.~\ref{fig:dispersion_incorrect}) display the opposite trajectory: they begin in a sub-Poisson regime ($D{<}1$) dominated by repetitive short fragments (e.g., repeated ``th'' or ``yo''), then transition toward Poisson ($D{\approx}1$) as errors become sparse and independent. Post-transition, incorrect-word dispersion hovers near unity, consistent with random, uncorrelated residual noise. The sub-Poisson $\to$ Poisson shift for incorrect words mirrors the near-Poisson $\to$ sub-Poisson shift for correct words, revealing a coordinated reorganization.

The coordination between dispersion and KL divergence is particularly revealing. While dispersion directly measures variance relative to the mean, KL divergence captures the full shape of the distribution. During the transition, both metrics show cusps, but they detect slightly different features: dispersion tracks the \emph{regime change} (Poisson $\to$ sub-Poisson for correct words), while KL tracks the \emph{magnitude of structural deviation}. The fact that both exhibit synchronized discontinuities in the same narrow epoch window—without being mathematically redundant—provides robust evidence that the transition reflects a genuine internal reorganization rather than an artifact of a single metric.

\FloatBarrier

\subsection{Summary}
Across all five seeds, the training trajectory divides naturally into three regimes—before, during, and after a discontinuity centered around steps 230--250.

\textbf{Before the discontinuity} (steps $\lesssim$230), the model remains in an exploratory regime. Generated text consists mostly of short, fragmented tokens; both dispersion and KL divergence indicate near-Poisson behavior for correct words ($D{\approx}1$, low KL); and the incorrect vocabulary slowly expands. Correct words appear sporadically and are typically one or two characters long, indicating limited compositional structure.

\textbf{During the discontinuity} (steps $\sim$230--250), every external metric reorganizes simultaneously. Both dispersion and KL divergence exhibit sharp, synchronized cusps—dispersion temporarily returns to Poisson while KL divergence peaks, indicating maximum distributional deviation. The incorrect vocabulary reaches its maximum diversity, and the model temporarily produces more errors as it experiments with new token combinations. Average correct-word length rises sharply in an S-shaped jump from $\sim$1.5$\,\to\,$2.5 characters, and prefix tracking (e.g., for \emph{``you''}) shows isolated fragments fusing into complete, stable words. This brief phase marks a concentrated restructuring of internal representations, where exploratory variability and combinatorial mixing peak.

\textbf{After the discontinuity} (steps $\gtrsim$250), exploratory variability subsides. The incorrect vocabulary collapses while the correct vocabulary grows steadily; dispersion drops below unity ($D{<}1$) and stabilizes, while KL divergence for correct words remains elevated (reflecting structured, non-Poisson regularity) and KL for incorrect words approaches zero (sparse, independent errors). Generated outputs become longer, coherent, and linguistically stable, reflecting consolidation of learned structure.

Crucially, the synchronization across \emph{multiple independent metrics}—dispersion (variance structure), KL divergence (full distributional form), word length (linguistic complexity), and vocabulary dynamics (compositional reorganization)—provides converging evidence that this is a genuine phase-transition-like reorganization. No single metric alone would be as convincing; their collective cusp in the same narrow epoch window rules out metric artifacts and supports the interpretation of an abrupt internal restructuring.

\section{Discussion}

\subsection{Poisson-Centered Diagnostics and Detection in Linear Training Space}
Our study addresses three central questions about phase transitions in language models. The second question—whether such transitions can be detected directly in linear training space rather than only after log rescaling—is answered affirmatively through our Poisson-centered diagnostic framework. By tracking both dispersion and KL divergence alongside vocabulary statistics, we observe synchronized cusp-like discontinuities that occur directly along the raw epoch axis, without logarithmic transformation of training steps or compute. This demonstrates that emergent reorganizations are intrinsic properties of the learning process, not artifacts of visualization conventions or metric rescaling.

Before the transition, incorrect words display variance suppression ($D < 1$), forming narrow sub-Poisson distributions dominated by repetitive fragments such as ``th'' or ``yo.'' Correct words appear sparsely and follow near-Poisson randomness, with low KL divergence from the baseline. At the transition, all metrics converge within a narrow critical band: correct-word dispersion temporarily approaches Poisson ($D \!\approx\! 1$) while KL divergence peaks sharply, indicating that although the variance ratio returns to unity, the full distribution shape deviates maximally from true Poisson. This divergence between dispersion and KL during the transition reveals complementary information: dispersion tracks the \emph{variance regime}, while KL captures the \emph{full distributional structure}. The fact that both exhibit cusps—but measure different statistical properties—strengthens the case for genuine reorganization rather than a metric artifact.

Afterward, the incorrect vocabulary collapses while correct-word usage stabilizes in a sub-Poisson regime (low $D$, elevated KL reflecting structured regularity), and average word length increases. This coordinated Poisson $\to$ sub-Poisson swap across multiple independent probes provides a clear statistical signature of the transition, directly answering our second research question: phase-transition-like discontinuities can be detected in linear training space through Poissonian diagnostics.

\subsection{Early Lexical Reorganization as the Transition Mechanism}
The observed transition is not a gradual refinement of accuracy but a discrete reorganization in the model's word-generation strategy. Evidence across multiple linguistic indicators reveals that the transition is driven by the abrupt emergence of longer, coherent words: 
(i) the mean correct-word length shows an S-shaped rise from $\sim$1.5 to $\sim$2.5 characters, representing a jump in linguistic complexity; 
(ii) prefix tracking of the word ``you'' reveals consolidation, where single-letter (\emph{y}) and two-letter (\emph{yo}) fragments merge into the stable three-letter form (\emph{you}) precisely at the transition epoch; and 
(iii) the unique incorrect vocabulary, which expands during fragment proliferation, declines sharply as fragments are absorbed into valid words. 

These convergent signatures reveal a qualitative reorganization from fragmentary to compositional word formation. Crucially, this shift occurs \emph{early} in training—around epochs 230--250—well before training or validation loss stabilizes. The model achieves lexical coherence not through late-stage fine-tuning but through an abrupt mid-training reorganization. This directly answers the third research question: phase-transition-like reorganizations can emerge at early stages of training, marking the onset of lexical coherence rather than late-stage convergence. The synchronized timing across dispersion cusps, KL divergence peaks, vocabulary dynamics, and word-length jumps confirms that these changes reflect a single, coordinated restructuring event.

\subsection{Temporary Degradation as Evidence of Barrier-Crossing Dynamics}

A particularly striking feature of the transition—and one that distinguishes genuine phase-transition-like reorganization from gradual learning—is that multiple metrics indicate \emph{temporary performance degradation} before improvement. The incorrect vocabulary reaches its maximum diversity at step 250 (Fig.~\ref{fig:uniq_incorrect}), meaning the model generates \emph{more} errors during the transition than in the preceding 100 epochs. Simultaneously, correct-word dispersion reverts toward Poisson ($D{\approx}1$) from its earlier sub-Poisson tendency, indicating that the weak regularity established early in training is temporarily disrupted. KL divergence also peaks sharply during this window, confirming maximum distributional instability. The model does not improve monotonically; it briefly becomes \emph{worse} before reorganizing into a more coherent state.

This ``destabilization before consolidation'' pattern is diagnostic of first-order phase transitions in statistical mechanics, where systems must overcome nucleation barriers to transition between competing stable states. In the framework developed by Rubin, Seroussi, and Ringel for grokking, such behavior arises when an effective potential has two minima—one corresponding to memorization and another to generalization—separated by a barrier. The system must temporarily increase its effective "energy" (disorder, entropy, error rate) to escape the memorization basin and reach the generalization basin. During the barrier-crossing phase, the system exhibits mixed-phase behavior: neither fully in the old regime nor fully in the new, leading to heightened variability and apparent regression.

Our linguistic data provide empirical evidence for this mechanism in a character-level language model. Fragment proliferation—the temporary surge in incorrect vocabulary—is not wasted computation but a necessary exploratory phase. The model must generate and test many erroneous combinations (``th,'' ``yo,'' ``ou,'' etc.) before discovering which fragments compose into valid words. This exploration is visible as increased error diversity and loss of early regularity. Once sufficient fragments have been tested, the system consolidates: incorrect vocabulary collapses as fragments are absorbed into multi-character words (e.g., ``y'' + ``o'' $\to$ ``yo,'' then ``yo'' + ``u'' $\to$ ``you''), and correct-word usage stabilizes into sub-Poisson regularity. The transition is abrupt not because learning suddenly accelerates, but because the system commits to a compositional strategy after exhausting fragmentary exploration.

This finding directly parallels grokking, where models trained on algorithmic tasks exhibit prolonged plateaus in generalization accuracy—sometimes hundreds of thousands of steps—before sudden jumps to near-perfect performance. Power et al.\ and Agarwal et al.\ documented that during these plateaus, internal representations are reorganizing even though external accuracy remains flat or fluctuates. Our contribution extends this phenomenon to linguistic tasks and provides statistical signatures—vocabulary proliferation, dispersion reversion, KL divergence peaks—that \emph{predict} impending reorganization. Unlike grokking studies that focus on memorization-to-generalization transitions in supervised settings, we observe exploration-to-consolidation transitions in unsupervised next-token prediction, suggesting that barrier-crossing dynamics are a general feature of neural training.

The broader implication is that \emph{stagnation or temporary degradation during training may signal impending breakthrough rather than failure}. Standard training practices often interpret validation plateaus or upticks in error rate as signs to stop training, adjust hyperparameters, or discard the model. Our findings suggest an alternative interpretation: such periods may reflect productive exploration necessary for qualitative reorganization. A model generating more errors at step 250 than step 200 is not necessarily failing—it may be sampling the combinatorial space required to discover compositional structure. Recognizing this pattern could inform more robust training protocols that tolerate or even encourage exploration phases, particularly in settings where emergent capabilities (reasoning, multi-step inference, compositionality) require crossing representational barriers.

Future work should investigate whether similar exploration-before-consolidation dynamics precede higher-level emergent abilities in larger models. If sentence-level or paragraph-level coherence also emerges through temporary destabilization and vocabulary churn, then Poisson-based diagnostics could serve as early-warning indicators of impending capability jumps. Conversely, understanding the conditions under which models successfully traverse barriers versus becoming trapped in suboptimal basins could guide interventions—learning rate schedules, regularization adjustments, curriculum design—that facilitate rather than suppress exploration. Our small-model study provides proof of principle that such dynamics are detectable and interpretable through statistical-mechanical frameworks, opening a path toward more principled management of the nonlinear, non-monotonic nature of neural learning.

\subsection{Observability in Small Models and Parallels with Physical Phase Transitions}
In physical systems, phase transitions occur when order parameters change discontinuously as control variables such as temperature or pressure cross a critical threshold. Analogously, in our model, the control variable is training epoch, while linguistic statistics—dispersion, KL divergence, vocabulary count, and word length—serve as order parameters. All exhibit coordinated discontinuities in the same narrow epoch range (steps 230--250), consistent with a critical reorganization of the generative regime.

The use of \emph{multiple, independent order parameters} is crucial. In physics, genuine phase transitions are characterized by the simultaneous reorganization of multiple observables—not just one. Here, dispersion and KL divergence measure fundamentally different aspects of the count distributions (variance structure vs.\ full distributional form), yet both display synchronized cusps. This convergence across non-redundant metrics mirrors how physical phase transitions are confirmed through multiple experimental probes (e.g., heat capacity, magnetization, correlation length), each sensitive to different aspects of the critical reorganization.

This behavior parallels two known phenomena: \emph{grokking}, where small models abruptly transition from memorization to generalization on algorithmic tasks, and emergent abilities in large language models, where new capabilities appear suddenly beyond critical scale thresholds. By observing these reorganizations in a 3.6M-parameter transformer trained on a modest corpus, we directly address the first research question: phase-transition-like behavior is \emph{not} exclusive to large-scale systems or massive models observed through log-scaled compute. Such discontinuities can manifest in small, tractable models when analyzed with appropriate statistical probes. This reinforces the idea that emergent dynamics are scale-invariant in form—they represent fundamental properties of neural learning, not mere artifacts of size. Poisson-based diagnostics—combining both dispersion and KL divergence—thus provide a unifying framework for linking the nonlinear dynamics of learning to the physics of critical phenomena.

\subsection{Addressing the Schaeffer et al.\ Critique}
Our methodological approach directly addresses concerns raised by Schaeffer, Miranda, and Koyejo regarding the reality of emergent abilities. They demonstrated that many reported emergent capabilities arise from discrete, non-linear evaluation metrics rather than genuine phase transitions, and that continuous metrics often reveal smooth scaling instead of discontinuities. Our study heeds this critique by employing \emph{continuous, internal statistical measures}—dispersion and KL divergence—rather than binary task accuracy thresholds.

Critically, we detect synchronized discontinuities across \emph{multiple independent continuous metrics}, each measuring different statistical properties. Dispersion quantifies variance relative to the mean; KL divergence captures full distributional deviation; word length measures linguistic complexity; vocabulary dynamics track compositional reorganization. No single metric alone would suffice to claim a phase transition, but their simultaneous cusps within the same narrow epoch window provide converging evidence that cannot be explained by metric artifacts. Furthermore, these signatures appear in raw training space without rescaling, and are invisible in standard loss curves, confirming that we are detecting internal reorganization rather than external scoring effects.

\subsection{Limitations and Scope}
While the findings provide convergent evidence for a small-model phase transition, several limitations warrant caution. 
First, we analyzed a single architecture (3.6M parameters) and a single dataset (Tiny Shakespeare); generalization to larger models, multilingual corpora, or instruction-tuned datasets remains untested. 
Second, character-level tokenization differs from subword or BPE segmentation used in practical LLMs, and the correct/incorrect heuristic may misclassify rare but valid words or neologisms. 
Third, our study focuses on \emph{external} statistical signatures—dispersion, KL divergence, vocabulary composition—while the correspondence between these metrics and internal representational shifts (such as activation clustering, attention alignment, or hidden-state geometry) remains open. 
Fourth, alternative decoding methods (top-$k$, nucleus sampling, beam search) might influence surface statistics and potentially shift the apparent transition region, though we expect the qualitative reorganization to persist. 
Fifth, we have not yet examined universality across model sizes, datasets, or architectural families; testing whether the same Poisson $\to$ sub-Poisson signature recurs under variation would strengthen claims of genuine criticality.
Finally, while we employ two complementary Poisson-based metrics—dispersion and KL divergence—future work should explore whether other statistical probes (e.g., higher-order cumulants, entropy measures, or correlation functions) reveal additional structure during the transition. The robustness of the synchronized cusps across dispersion and KL suggests the transition is genuine, but expanding the diagnostic toolkit would further strengthen this interpretation.

\section{Future Work}
Future research should extend these analyses along four complementary directions. 

\textbf{(1) Scaling and universality:} Test whether Poisson--sub-Poisson reorganizations recur across model scales (from small to billion-parameter systems), diverse datasets (multilingual corpora, code, instruction-tuned data), and architectural families (encoder-decoder, causal transformers, state-space models). Establish whether the transition epoch, dispersion dynamics, and KL divergence cusps exhibit universal patterns or depend systematically on scale and domain. This would assess whether our findings reflect a general property of language model training or are specific to small character-level systems.

\textbf{(2) Quantitative characterization:} Apply finite-size scaling analysis to distinguish genuine critical phenomena—characterized by power-law divergences and universal scaling exponents—from simple threshold effects. Estimate critical exponents for dispersion, KL divergence, and correlation lengths near the transition. Determine whether the transition is first-order (discontinuous order parameter, hysteresis, mixed phases) or continuous (critical slowing, diverging fluctuations), following the framework of Rubin, Seroussi, and Ringel. This would rigorously classify the transition within the taxonomy of statistical mechanics.

\textbf{(3) Mechanistic linkage:} Connect external Poisson-based statistics with internal network dynamics. Investigate whether dispersion cusps coincide with changes in hidden-state correlation lengths, attention-map reorganizations, embedding-space clustering, or gradient flow patterns. Use mechanistic interpretability techniques—such as singular value decomposition of representations, neuron activation analysis, and circuit tracing—to identify which internal structures reorganize during the transition. This would bridge behavioral signatures (dispersion, KL) with representational mechanisms, revealing \emph{how} the model reorganizes internally to produce the observed statistical shift.

\textbf{(4) Higher-level transitions:} Investigate whether similar Poisson--sub-Poisson discontinuities appear at sentence or paragraph levels, signaling the sudden emergence of grammatical coherence, semantic consistency, or discourse structure. Extend the windowing framework to multi-word sequences and track dispersion/KL divergence for syntactic constructions (e.g., subject-verb agreement, clause embedding). Test whether linguistic hierarchies exhibit cascading transitions—lexical coherence first, then syntactic, then semantic—each marked by distinct reorganization signatures.

Together, these efforts would test whether Poisson-centered metrics provide a general signature of emergent coherence across linguistic hierarchies, model scales, and training regimes.

Taken together, these findings answer the three motivating questions posed in the introduction with converging evidence: 
(1) phase-transition-like reorganizations are \emph{not} unique to large models but occur in small transformers when probed with appropriate statistical metrics; 
(2) they can be detected directly in \emph{linear training space} without logarithmic rescaling, using continuous Poisson-based diagnostics rather than binary task thresholds; and 
(3) they emerge at \emph{early stages of training}—around epochs 230--250 in our 3.6M-parameter model—as lexical coherence forms, well before loss convergence. 
The synchronization of multiple independent metrics—dispersion, KL divergence, word length, and vocabulary dynamics—within the same narrow epoch window provides robust evidence that these phenomena reflect genuine internal reorganizations, not metric artifacts or gradual drift. 
These results collectively demonstrate that emergent phenomena are intrinsic, measurable properties of neural language model training dynamics, observable even at modest scale.

\section{Conclusion}
We have presented converging evidence that a small transformer undergoes a distinct, phase-transition-like reorganization during training, detectable directly in linear training space through Poisson-centered diagnostics. Both dispersion and KL divergence reveal synchronized cusps at epochs 230--250, where correct-word statistics briefly return toward Poissonian behavior before settling into sub-Poisson regularity. Unique incorrect vocabulary first expands during fragment proliferation, then collapses as errors are abandoned, while the unique correct vocabulary grows monotonically. Average correct-word length exhibits a sharp S-curve increase from $\sim$1.5 to $\sim$2.5 characters, and prefix tracking of the word ``you'' illustrates how single- and two-letter fragments consolidate into a stable three-letter form precisely at the transition epoch.

Together, these findings demonstrate that emergent-like reorganizations can arise even in modest transformer models at early stages of training, without requiring log-scaling or massive compute. By employing multiple independent, continuous statistical probes—dispersion (variance structure), KL divergence (full distributional form), word length (linguistic complexity), and vocabulary dynamics (compositional reorganization)—we establish a quantitative framework for detecting discontinuous learning dynamics that avoids the metric artifacts identified by recent critiques of emergent abilities. The synchronization of cusps across non-redundant metrics provides robust evidence for genuine internal restructuring.

This approach bridges concepts from statistical physics with linguistic emergence, showing that coherence at higher linguistic levels—words, sentences, and beyond—may likewise arise through abrupt structural reorganizations rather than gradual accumulation. Small models thus serve as powerful laboratories for studying criticality in neural systems, and Poisson-based diagnostics provide a unifying lens for understanding the nonlinear dynamics of learning in language models. Our results suggest that phase-transition-like phenomena are fundamental properties of neural training, observable at any scale when appropriate probes reveal the underlying reorganization.

\EOD
\end{document}